\PassOptionsToPackage{table}{xcolor}

\documentclass[sigconf, nonacm]{acmart}
\usepackage{multirow}
\usepackage{enumitem}
\usepackage{xcolor}
\usepackage{colortbl}
\usepackage{graphicx}      
\usepackage{svg}
\usepackage{subcaption}    
\definecolor{lightgray}{gray}{0.9}
\definecolor{revgreen}{RGB}{0,100,0}
\usepackage{amsfonts}
\usepackage{microtype}
\usepackage{array}

\usepackage{tikz} 

\newcommand\vldbdoi{XX.XX/XXX.XX}
\newcommand\vldbpages{XXX-XXX}
\newcommand\vldbvolume{20}
\newcommand\vldbissue{2}
\newcommand\vldbyear{2026}
\newcommand\vldbauthors{\authors}
\newcommand\vldbtitle{\shorttitle} 
\newcommand\vldbavailabilityurl{https://github.com/ldp2211479/OntoTKGE}
\newcommand\vldbpagestyle{empty} 

\begin{document}
\title{OntoTKGE: Ontology-Enhanced Temporal Knowledge Graph Extrapolation}







\author{Dongying Lin$^{1*}$, Yinan Liu$^{1*}$, Shengwei Tang$^{1}$, Bin Wang$^1$, Xiaochun Yang$^{1\dagger}$}
\affiliation{%
  \institution{$^1$Northeastern University, Shenyang, China}
}
\email{2472128@stu.neu.edu.cn, liuyinan@cse.neu.edu.cn, tangshengwei40@gmail.com, {binwang,yangxc}@mail.neu.edu.cn}

\begin{abstract}
Temporal knowledge graph (TKG) extrapolation is an important task that aims to predict future facts through historical interaction information within KG snapshots. A key challenge for most existing TKG extrapolation models is handling entities with sparse historical interaction. The ontological knowledge is beneficial for alleviating this sparsity issue by enabling these entities to inherit behavioral patterns from other entities with the same concept, which is ignored by previous studies. In this paper, we propose a novel encoder-decoder framework OntoTKGE that leverages the ontological knowledge from the ontology-view KG (i.e., a KG modeling hierarchical relations among abstract concepts as well as the connections between concepts and entities) to guide the TKG extrapolation model's learning process through the effective integration of the ontological and temporal knowledge, thereby enhancing entity embeddings. OntoTKGE is flexible enough to adapt to many  TKG extrapolation models.
Extensive experiments on five data sets demonstrate that OntoTKGE not only significantly improves the performance of many TKG extrapolation models but also surpasses many state-of-the-art (SOTA) baseline methods. 
\end{abstract}

\maketitle

\begingroup
\renewcommand{\thefootnote}{}
\footnotetext{\noindent $^*$Equal contributions. $^\dagger$Corresponding author.}
\endgroup

\pagestyle{\vldbpagestyle}
\begingroup\small\noindent\raggedright\textbf{PVLDB Reference Format:}\\
\vldbauthors. \vldbtitle. PVLDB, \vldbvolume(\vldbissue): \vldbpages, \vldbyear.\\
\href{https://doi.org/\vldbdoi}{doi:\vldbdoi}
\endgroup
\begingroup
\renewcommand\thefootnote{}\footnote{
\noindent This work is licensed under the Creative Commons BY-NC-ND 4.0 International License. Visit \url{https://creativecommons.org/licenses/by-nc-nd/4.0/} to view a copy of this license. For any use beyond those covered by this license, obtain permission by emailing \href{mailto:info@vldb.org}{info@vldb.org}. Copyright is held by the owner/author(s). Publication rights licensed to the VLDB Endowment. \\
\raggedright Proceedings of the VLDB Endowment, Vol. \vldbvolume, No. \vldbissue\ %
ISSN 2150-8097. \\
\href{https://doi.org/\vldbdoi}{doi:\vldbdoi} \\
}\addtocounter{footnote}{-1}\endgroup

\ifdefempty{\vldbavailabilityurl}{}{
\vspace{.3cm}
\begingroup\small\noindent\raggedright\textbf{PVLDB Artifact Availability:}\\
The source code, data, and/or other artifacts have been made available at \url{\vldbavailabilityurl}.
\endgroup
}

\section{Introduction}
 Temporal knowledge graphs (TKGs) \cite{sigmod1,hisres,booter} represent knowledge as quadruples in the form of (\textit{subject}, \textit{relation}, \textit{object}, \textit{timestamp}), extending static KGs \cite{Liu4} representing knowledge as facts by introducing the temporal attributes to capture the dynamic knowledge evolution over time. The task of TKG reasoning aims to infer new knowledge from historical information within TKGs and is generally divided into two settings: interpolation \cite{booter} and extrapolation \cite{renet}. The former aims to complete missing facts in history, while the latter aims to forecast unknown facts that will occur in the future. 
 This work focuses on the extrapolation setting, with applications like medical aided diagnosis system \cite{diagnosis_system_v1} and event prediction \cite{event_prediction,event_prediction2}.

\begin{figure}[t]
  \centering
  \includegraphics[width=0.99\linewidth]{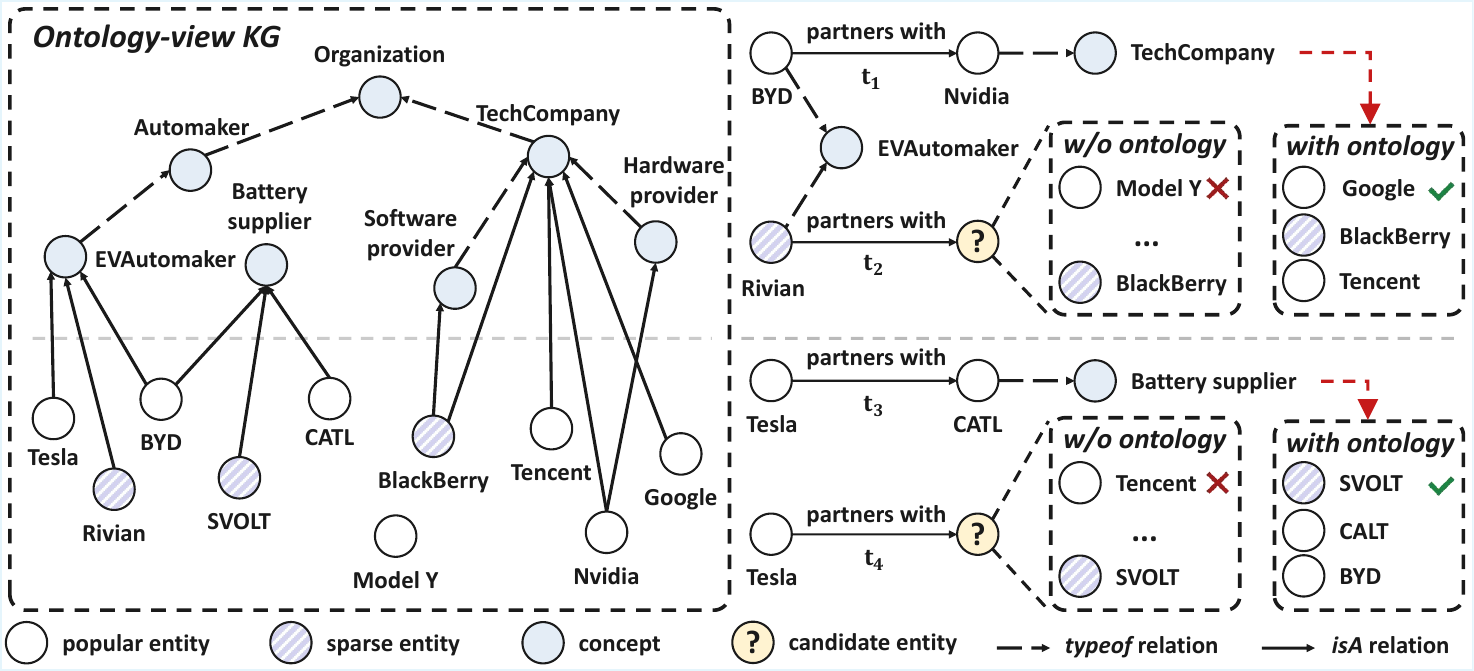}
  \vspace{-4mm}
  \caption{An illustration of the ontology-view KG.}
  \vspace{-6.8mm}
  \label{fig:ontology enhance tkg}
\end{figure}

Currently, a multitude of TKG extrapolation models (e.g., RE-GCN \cite{regcn}, TiRGN \cite{tirgn}, LogCL \cite{logcl}, and HisRES \cite{hisres}) have been proposed. 
Despite their effectiveness, they often struggle to predict sparse entities (i.e., entities with few or no historical interaction data). To address this issue, recent studies (e.g., LLM-DA \cite{llmda} and ANEL \cite{anel}) enhance TKG extrapolation models based on rules adapted by large language models (LLMs) \cite{llmda} or latent semantic relationships derived from neighborhood structures \cite{anel}.
However, for sparse entities, ANEL may identify incorrect latent neighbors due to unreliable embeddings, while LLM-DA often fails to obtain high-confidence logical rules due to the scarcity of historical paths for rule extraction.
In this paper, our task is to design an enhancement framework to address the above inherent limitations.

Intuitively, an ontology-view KG, which models hierarchical relations among concepts and the connections between concepts and entities, is beneficial for enhancing entity embeddings. This is achieved by enabling entities with sparse historical interactions to inherit behavioral patterns from popular entities (i.e., entities with rich historical interaction data) that share the same concepts.
As shown in Figure \ref{fig:ontology enhance tkg}, the benefits are twofold. First, when forecasting the future behavior of the sparse entity \textit{``Rivian''}, its limited historical interaction data hinders TKG extrapolation models from identifying suitable candidates. By associating \textit{``Rivian''} with its concept \textit{``EVAutomaker''}, it can inherit the behavioral patterns from popular entities in the same category like \textit{``BYD''}, which partnered with \textit{``Nvidia''}. This enables the model to infer that potential partners lie within the concept \textit{``TechCompany''}, which \textit{``Nvidia''} belongs to, thus guiding the candidate search effectively. 
Second, existing TKG extrapolation models may fail to regard a sparse entity as a valid candidate, even if it is the correct answer. For example, when predicting a future battery supplier for \textit{``Tesla''}, the model may rely on historical interaction data involving the popular entity \textit{``CATL''} (an entity within the concept \textit{``Battery Supplier''}) and overlook the sparse but plausible candidate \textit{``SVOLT''}. The ontology-view KG solves this by explicitly classifying \textit{``SVOLT''} as a \textit{``Battery Supplier''}, keeping such relevant yet sparse entities in the candidate set and capturing the actual entity. Hence, how to effectively use the ontological knowledge to enhance the embeddings of TKG extrapolation models is crucial, and requires further exploration.

Ontologies are commonly used to enhance static KG embeddings \cite{joie,ciss,hypercl}. They typically learn entity and concept embeddings separately, and then employ alignment losses to constrain entity embeddings with ontological knowledge. An intuitive way to apply these methods to TKGs is to obtain static ontology embeddings independently and then use similar alignment losses to constrain the dynamic entity embeddings generated by TKG extrapolation models. However, our experiments reveal that this direct migration is often ineffective. 
A potential explanation is that learning the ontology embeddings separately from the model neglects the benefits of incorporating the ontological knowledge from the ontology-view KG to guide the model's learning process.
Furthermore, applying a uniform ontological constraint to all entities may hinder the model from capturing their distinct evolutionary patterns.

To address these issues, we propose a novel enhancement framework \textbf{OntoTKGE} that integrates the \underline{\textbf{Onto}}\-log\-i\-cal and temporal knowledge to improve the performance of existing \underline{\textbf{TKG}} \underline{\textbf{E}}x\-trap\-o\-la\-tion models via an encoder-decoder architecture. First, OntoTKGE automatically constructs an ontology-view KG for any TKG through a hybrid method combining the LLM with entity linking and retrieval techniques. Next, OntoTKGE generates entity embeddings within TKGs via a global ontology-aware evolutionary encoder, which incorporates the ontological knowledge to initialize entity embeddings via hierarchical entailment learning and uses the TKG extrapolation model's encoder to evolve these embeddings across a sequence of KG snapshots. To alleviate the loss of ontological knowledge across KG snapshots during the above process, OntoTKGE generates supplementary embeddings specifically for query entities via a local ontology-aware relevance encoder based on a simple graph convolutional network. 
Subsequently, 
OntoTKGE uses a contrastive-enhanced gated fusion unit. It adaptively fuses embeddings from different encoders via a gated fusion unit, and regularizes them by minimizing the distance between embeddings of the same entity from different encoders while maximizing the distance between those of different entities, drawing inspiration from contrastive learning.
Finally, the existing TKG extrapolation model's decoder can be applied to the fused embeddings for the final prediction. 

%
%


Our main contributions are summarized as follows: 
(1) To the best of our knowledge, this is the first work to explore the use of the ontological knowledge for enhancing TKG extrapolation. 
(2) We propose a ontology-based framework OntoTKGE to enhance existing TKG extrapolation models. OntoTKGE is flexible enough to adapt to many TKG extrapolation models. 
(3) We construct Wiki35k, a new temporal data set for evaluating TKG extrapolation.
(4) Extensive experiments on five real-world data sets demonstrate that OntoTKGE not only effectively improves the performance of many SOTA TKG extrapolation models but also outperforms all baselines.

\vspace{-2mm}
\section{Notations and Definitions}
In this paper, a TKG is denoted by a sequence of KG snapshots, i.e., ${G}= \{\mathcal{G}_1,\mathcal{G}_2, \dots, \mathcal{G}_{|T|}\}$. Each KG snapshot $\mathcal{G}_t=(\mathcal{E}, \mathcal{R}, \mathcal{F}_I^t)$ is a directed multi-relational graph, denoting the set of facts that occur at a specific timestamp $t \in T$. $\mathcal{E}$ and $\mathcal{R}$ are entity and relation sets, respectively. $\mathcal{F}_I^t$ is the set of facts observed at $t$ ($I$ denotes the instance view). Each fact at $t$ is denoted by a quadruple $(s, r_I, o, t)$, where $s \in \mathcal{E}$ is the subject entity, $o \in \mathcal{E}$ is the object entity, and $r_I \in \mathcal{R}$ is the relation between $s$ and $o$. Given historical KG snapshots $\{\mathcal{G}_1,\mathcal{G}_2,\dots,\mathcal{G}_{t}\}$ and a query $(s,r_I,?,t+1)$ (resp. $(?,r_I,o,t+1)$), our task is to predict the object entity (resp. subject entity).


\vspace{-2mm}
\section{The Framework OntoTKGE}
The overall framework OntoTKGE is shown in Figure \ref{fig:overview_framework}. 

\vspace{-2mm}
\subsection{Ontology-View KG Construction} \label{subsec:3.1}
Due to the lack of high-quality ontologies within data sets of the TKG extrapolation task, we propose a fully automated ontology construction pipeline for the TKG based on a hybrid method that integrates the LLM \cite{gpt4} with the conventional entity linking and retrieval methods. 
For each entity $e \in \mathcal{E}$, we aggregate its associated historical facts $\overline{\mathcal{F}}_e=\{(s,r,o,t) \in \mathcal{F}^t_I\ |\ s=e\ \vee \ o=e\}$ and prompt the LLM to generate a description and disambiguated name as the context, while using original names for entities which do not have associated facts~\cite{Liu1,Liu2}. We then link $e$ to the corresponding entity in a static KG by querying the Wikidata API and using ReFinED \cite{refined,Liu3}, an end-to-end entity linking model, to retrieve candidate entities and utilize the LLM to select the entity with the highest probability from candidate entities.
For each successfully linked entity, we retrieve its three-hop neighborhoods in the static KG via SPARQL queries, selecting relations that represent entity categories and static properties, including \textit{instance of}, \textit{occupation}, and \textit{country}, and hierarchical relations among concepts, including \textit{subclass of} to collect the associated concepts.
For unlinked entities, we perform an entity typing strategy that begins with a coarse-grained retrieval of concept candidates using a retrieval model, followed by a fine-grained ranking via the LLM to assign the most suitable concepts and relations. 
Specifically, the retrieval model first retrieves concept candidates with the entity name and corresponding description as the query, and the LLM ranks them to select up to six concepts. After concept selection, we assign the most frequent relation of the selected concept in existing facts between entities and concepts. For instance, if most linked entities are connected to \textit{politician} by \textit{occupation}, the same relation is assigned to an unlinked entity typed as \textit{politician}.
Thus, we can construct the generated ontology-view KG denoted by $G_O=(\mathcal{E}_O,\mathcal{R}_O,\mathcal{F}_O)$, where the subscript $O$ denotes the ontology view. $\mathcal{E}_O$ is the union of the set of entities $\mathcal{E}$ and the set of concepts $\mathcal{C}$. $\mathcal{R}_O$ denotes the set of relations. $\mathcal{F}_O$ consists of facts that connect entities with concepts and facts that connect concepts with concepts through hierarchical relations, providing a unified concept and relation vocabulary for linked and unlinked entities. Each ontological fact is denoted by $(ec, r_O, c)$, where $ec \in \mathcal{E}_O$ denotes an entity $e \in \mathcal{E}$ or a concept $c \in \mathcal{C}$, and $r_O \in \mathcal{R}_O$ is the relation between $ec$ and $c$.


\subsection{Global Ontology-Aware Evolutionary Encoder} \label{subsec:global encoder}
This encoder uses the prior ontological knowledge from the ontology-view KG ${G}_O$ to initialize embeddings of all entities in the TKG and use the TKG extrapolation model's encoder to evolve these entity embeddings across multiple KG snapshots. Specifically, it uses composition-based multi-relational graph convolutional network (\textit{CompGCN}) \cite{compgcn} to capture the interactions between entities and concepts within ${G}_O$. 
The \textit{CompGCN}'s update process can be described as follows:
\begin{equation}
\resizebox{0.98\hsize}{!}{$ 
\begin{aligned}
\textbf{h}_{g,c}^{(j+1)}&=\mathit{CompGCN}_{\text{global}}(\textbf{h}_{g,ec}^{(j)},\textbf{h}_{g,r_O}^{(j)},\textbf{h}_{g,c}^{(j)}) \\
         &=\sigma_1(\frac{1}{v_c} \sum_{\substack{(ec,r_O), \\ \exists(ec,r_O,c)\in\mathcal{F}_{O}}} W_{1}^{(j)}f(\textbf{h}_{g,ec}^{(j)},\textbf{h}_{g,r_O}^{(j)})+W_{2}^{(j)}\textbf{h}_{g,c}^{(j)}),
\end{aligned}
$}
\end{equation}
where $\textbf{h}_{g,ec}^{(j)}$, $\textbf{h}_{g,c}^{(j)}$, and $\textbf{h}_{g,r_O}^{(j)}$ denote the embeddings of $ec$, $c$, and $r_O$ in the $j$-th layer of the \textit{CompGCN} in $G_O$, respectively. $\sigma_1$ is the RReLU activation function. $W_1^{(j)}$ and $W_2^{(j)}$ denote the feed-forward aggregation matrix and the self-connection matrix in the $j$-th layer, respectively. $v_c$ is a normalization constant that equals the in-degree of $c$. 
%
%
$f(\textbf{h}_{g,ec}^{(j)},\textbf{h}_{g,r_O}^{(j)})$ denotes the composition operation between the embedding of $ec$ and the relation embedding of $r_O$, including subtraction (\textit{Sub}) from TransE \cite{transe}, multiplication (\textit{Mult}) from DistMult \cite{distmult}, and circular-correlation (\textit{Corr}) from HolE \cite{hole}, defined as
    $f(\mathbf{h}_{g,ec}^{(j)},\mathbf{h}_{g,r_O}^{(j)})=\mathbf{h}_{g,ec}^{(j)}-\mathbf{h}_{g,r_O}^{(j)}$,
    $f(\mathbf{h}_{g,ec}^{(j)},\mathbf{h}_{g,r_O}^{(j)})=\mathbf{h}_{g,ec}^{(j)}\cdot \mathbf{h}_{g,r_O}^{(j)}$, and
    $f(\mathbf{h}_{g,ec}^{(j)},\mathbf{h}_{g,r_O}^{(j)})=\mathbf{h}_{g,ec}^{(j)} \star \mathbf{h}_{g,r_O}^{(j)}$, respectively.
More specifically, \textit{Sub} models relations as translations in the embedding space, \textit{Mult} uses element-wise multiplication to capture relational semantics, and \textit{Corr} uses circular correlation to create compositional representations and capture rich interactions.


Traditional \textit{CompGCN} transforms the relation embeddings from the previous layer to the next layer using a shared weighted matrix. To learn more diverse relational semantics at different hierarchical levels, we modify traditional \textit{CompGCN} by using independent and learnable vectors for relation embeddings at each layer. This operation is useful, which has been verified in our experiments (see Sec. \ref{subsubsec:variants}). For simplicity, we extract entity-specific embeddings that integrate ontological knowledge from the final layer of the output of \textit{CompGCN}, denoted by $\textbf{H}_{g} = (\textbf{h}_{g,e_1}, \textbf{h}_{g,e_2}, \dots, \textbf{h}_{g,e_{|\mathcal{E}|}})^T\ \in \mathbb{R}^{|\mathcal{E}| \times d}$, while the concept embeddings are denoted by $(\textbf{h}_{g,c_1}, \textbf{h}_{g,c_2}, \dots, \textbf{h}_{g,c_{|\mathcal{C}|}})^T$.
By feeding $\textbf{H}_{g}$ into the TKG extrapolation model's encoder at the first timestamp, we can obtain each entity's embedding at timestamp $t+1$. This process can be denoted as follows:
\begin{equation}
(\textbf{Z}_{t+1},\textbf{R}_{t+1})=\text{Encoder}(\{\mathcal{G}_1,\mathcal{G}_2,\dots,\mathcal{G}_{t}\}, \textbf{H}_{g}),
\end{equation}
where $\text{Encoder}$ is the encoder of an existing TKG extrapolation model (e.g., RE-GCN, TiRGN, and RETIA). $\textbf{Z}_{t+1} = (\textbf{z}_{t+1,e_1}, \textbf{z}_{t+1,e_2}, \dots, \\\textbf{z}_{t+1,e_{|\mathcal{E}|}})^T$ denotes entity embeddings at $t+1$.
$\textbf{R}_{t+1} = (\textbf{r}_{t+1,r_{I_1}}, \textbf{r}_{t+1,r_{I_2}},\\ \dots, \textbf{r}_{t+1,r_{I_{|\mathcal{R}|}}})^T$ denotes relation embeddings at $t+1$.

\begin{figure}[t]
  \centering
  \begin{tikzpicture}
    \node[
      inner sep=0pt,
      outer sep=0pt
    ] (image) {
      \includegraphics[
        width=0.99\linewidth
      ]{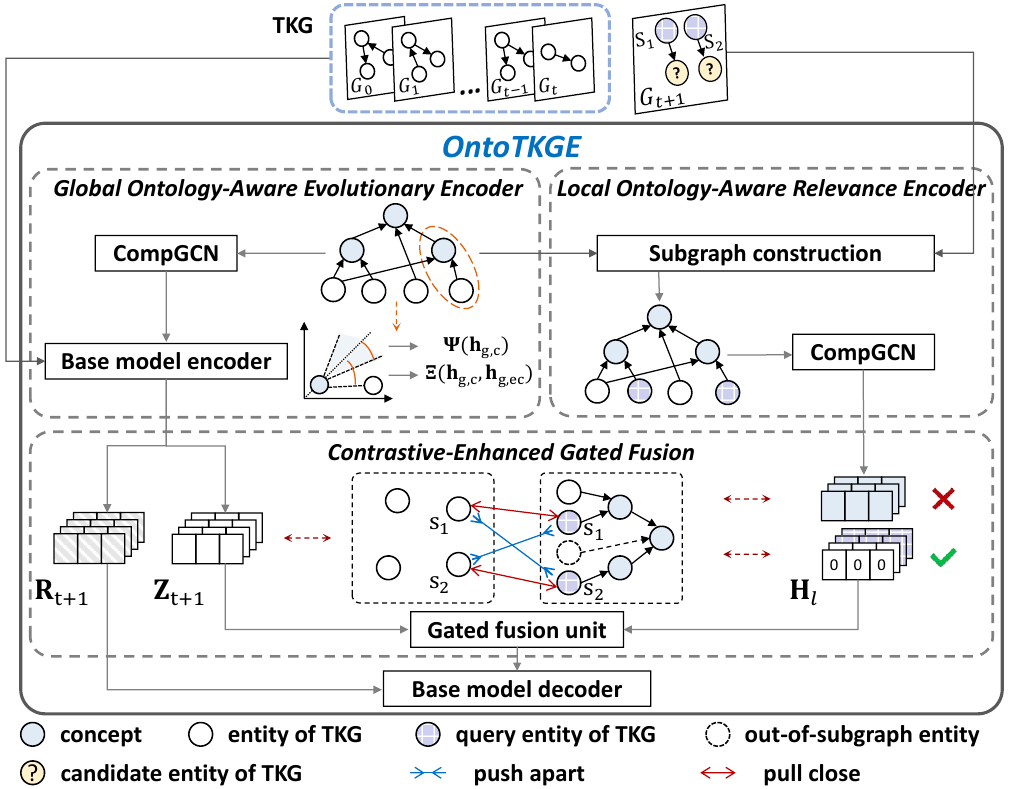}
    };
  \end{tikzpicture}
  \vspace{-4mm}
  \caption{Overview of our framework.}
  \vspace{-5mm}
  \label{fig:overview_framework}
\end{figure}


As illustrated in Figure \ref{fig:ontology enhance tkg}, the ontology-view KG exhibits a hierarchical tree structure, naturally aligning with entailment relations.
For instance, a specific concept in ontology-view KG like ``\textit{scientist}'' is inherently subsumed by a more general one, such as ``\textit{person}''.
To enhance embeddings furthermore, we introduce a hierarchical entailment constraint \cite{hierarchical-entailment-learning-18,hierarchical-entailment-learning-23} to the ontology-view KG, preserving its rich semantics better. 
Given the embeddings of a child $\textbf{h}_{g,ec}$ and its parent $\textbf{h}_{g,c}$ for an ontological fact, our optimization objective is to ensure that the embedding of each child node is geometrically contained within an \textit{entailment cone} defined by its parent. 
More specifically, the entailment cone of a parent node defines a region such that points in the region are semantically linked to the parent as its child concepts.
As shown in Figure \ref{fig:overview_framework}, we define the angle $\Xi(\textbf{h}_{g,c}, \textbf{h}_{g,ec})$ between $\textbf{h}_{g,c}$ and $(\textbf{h}_{g,ec} - \textbf{h}_{g,c})$ as follows:
\begin{equation}
    \resizebox{0.98\hsize}{!}{$ 
        \Xi(\textbf{h}_{g,c}, \textbf{h}_{g,ec}) = \cos^{-1}(\frac{\left\|\textbf{h}_{g,ec}\right\|^2 - \left\|\textbf{h}_{g,c}\right\|^2 - \left\|\textbf{h}_{g,c} - \textbf{h}_{g,ec}\right\|^2}{2 \left\|\textbf{h}_{g,c}\right\| \cdot \left\|\textbf{h}_{g,c}-\textbf{h}_{g,ec}\right\|}),
    $}
\end{equation}
where $\left\| \cdot \right\|$ denotes the Euclidean norm of a vector. 
The parent's entailment cone is defined by its half-aperture:
The aperture angle of the parent entailment cone is defined as $\Psi(\textbf{h}_{g,c}) = \sin^{-1}\left(\frac{K}{\|\textbf{h}_{g,c}\|}\right)$, where $K$ is a constant used for setting boundary conditions near the origin. A larger $K$ gives a wider parent entailment cone and allows more child embeddings to fall inside the cone, whereas a smaller $K$ gives a narrower cone and makes the entailment constraint stricter.
If $\Xi(\textbf{h}_{g,c}, \textbf{h}_{g,ec})$ is within $\Psi(\textbf{h}_{g,c})$, the partial order between $\textbf{h}_{g,c}$ and $\textbf{h}_{g,ec}$ is satisfied and no penalty is applied; otherwise, the deviation must be reduced. This is formulated via the loss function:
\begin{equation}
l_{\text{entail}}(\textbf{h}_{g,c}, \textbf{h}_{g,ec}) = \max(0, \Xi(\textbf{h}_{g,c}, \textbf{h}_{g,ec}) - \Psi(\textbf{h}_{g,c})).
\end{equation}
Finally, the total hierarchical entailment loss $\mathcal{L}_{\text{hie}}$ is the average loss over all pairs:
\begin{equation}
\mathcal{L}_{\text{hie}} = \frac{1}{|\mathcal{B}_{\text{hie}}|} \sum_{(ec, c) \in \mathcal{B}_{\text{hie}}} l_{\text{entail}}(\textbf{h}_{g,c}, \textbf{h}_{g,ec}),
\end{equation}
where $\mathcal{B}_{\text{hie}}=\{(ec,c)\ |\ (ec,r_O,c) \in \mathcal{F}_O\}.$

\subsection{Local Ontology-Aware Relevance Encoder} \label{subsec:local encoder}
Although the global ontology-aware evolutionary encoder incorporates the rich ontological prior knowledge into the initial entity embeddings, the static ontology knowledge may gradually attenuate as the temporal encoder learns from the KG snapshots.
For sparse entities, their embeddings are more prone to degradation due to limited historical facts for updates. To address this issue, we propose a local ontology-aware relevance encoder as an additional supplement of the global ontology-aware evolutionary encoder. Specifically, for each given query $(s,r_I,?,t+1)$, we first retrieve and construct an $N$-hop subgraph for $s$ from ${G}_O$, which can be denoted by $\hat{{G}}_O=\{\hat{\mathcal{E}}_O, \hat{\mathcal{R}}_O, \hat{\mathcal{F}}_O\} \subseteq {G}_O$. In this subgraph, $\hat{\mathcal{E}}_O \subseteq \mathcal{E}_O$ is the set of retrieved entities and concepts, and $\hat{\mathcal{R}}_O \subseteq \mathcal{R}_O$ is the set of retrieved relations. Each fact of $\hat{\mathcal{F}}_O \subseteq \mathcal{F}_O$ is represented as a triple $(\hat{ec}, \hat{r}_O, \hat{c})$, where $\hat{ec} \in \hat{\mathcal{E}}_O$ denotes an entity $\hat{e} \in \mathcal{E}$ or a concept $\hat{c} \in \mathcal{C}$, and $\hat{r}_O \in \hat{\mathcal{R}}_O$ is the relation between $\hat{ec}$ and $\hat{c}$. The subgraph contains the concepts and their hierarchical structure that are most directly related to $s$. To model the structural semantics of the subgraph, we employ another independent \textit{CompGCN} encoder:
\begin{equation}
    \textbf{h}_{l,\hat{c}}^{(j+1)}=\mathit{CompGCN}_{\text{local}}(\textbf{h}_{l,\hat{ec}}^{(j)},\textbf{h}_{l,\hat{r}_O}^{(j)},\textbf{h}_{l,\hat{c}}^{(j)}),
\end{equation}
where $\textbf{h}_{l,\hat{ec}}^{(j)}$, $\textbf{h}_{l,\hat{c}}^{(j)}$ and $\textbf{h}_{l,\hat{r}_O}^{(j)}$ denote the embeddings of $\hat{ec}$, $\hat{c}$, and $\hat{r}_O$ in the $j$-th layer of the \textit{CompGCN} in $\hat{G}_O$, respectively. Similarly, we extract the entity-specific embeddings that capture the ontological context of the subgraph from the final layer of the \textit{CompGCN} output and denote the matrix as $\textbf{H}_{l} = (\textbf{h}_{l,e_1}, \textbf{h}_{l,e_2}, \dots, \textbf{h}_{l,e_{|\mathcal{E}|}})^T \in \mathbb{R}^{|\mathcal{E}| \times d}$. The embeddings of entities in $\{\mathcal{E}_O - \hat{\mathcal{E}}_O\}$ are set to zero vectors. 

\vspace{-1mm}
\subsection{Contrastive-Enhanced Gated Fusion}
To combine the embeddings from the global ontology-aware evolutionary encoder and the local ontology-aware relevance encoder, we propose a gated fusion unit to obtain the overall representation:
\begin{equation}
    \begin{aligned}
    \hat{\textbf{Z}}_{t+1} &= \Theta \odot \textbf{H}_{l} + (1 - \Theta) \odot \textbf{Z}_{t+1}, 
    \end{aligned}
\end{equation}
where $\Theta = \sigma_2(W_3 \textbf{H}_{l} + W_4 \textbf{Z}_{t+1} + b)$, $W_3, W_4 \in \mathbb{R}^{d \times d}$ are learnable weight matrices, $b \in \mathbb{R}^d$ is a bias vector, $\sigma_2$ is the Sigmoid activation function, and $\odot$ denotes the Hadamard product. 

Note that the embeddings from the global ontology-aware evolutionary encoder and the embeddings from the local ontology-aware relevance encoder can be regarded as two distinct views that can characterize the same entity from different perspectives. 
Inspired by \cite{logcl}, we leverage contrastive learning to ensure semantic consistency between these two complementary views, enhancing the overall representation furthermore.
Specifically, let $\mathcal{M}_{t+1}$ denote a mini-batch that consists of entities in queries at timestamp $t+1$. For any given entity $u$ within $\mathcal{M}_{t+1}$, the embedding $\textbf{z}_{t+1,u}$ and embedding $\textbf{h}_{l,u}$ are considered as a positive pair as they refer to the same entity from two distinct views. 
Conversely, $\textbf{z}_{t+1,u}$ with the embedding $\textbf{h}_{l,j}$ of any other entity $j$ (where $j \neq u$) in the same mini-batch forms a negative pair. Formally, the contrastive loss $\mathcal{L}_{\text{cl}}$ is calculated as follows:
\begin{equation}
\mathcal{L}_{\text{cl}} = -\frac{1}{|\mathcal{M}_{t+1}|} \sum_{u=1}^{|\mathcal{M}_{t+1}|} \log \frac{e^{\text{sim}(\textbf{z}_{t+1,u},\textbf{h}_{l,u})/\tau}}{\sum_{j \neq u}^{|{\mathcal{M}_{t+1}}|} e^{\text{sim}(\textbf{z}_{t+1,u}, \textbf{h}_{l,j})/\tau}},
\end{equation}
where $\text{sim}(\cdot, \cdot)$ denotes the cosine similarity between two embeddings and $\tau$ is a temperature parameter. By minimizing $\mathcal{L}_{\text{cl}}$, the model is encouraged to learn a unified embedding space where the embeddings of the same entity from different encoders are aligned with each other, while separating those of different entities.

\vspace{-1mm}
\subsection{Prediction and Optimization}
Based on entity embeddings $\hat{\textbf{Z}}_{t+1} = (\hat{\textbf{z}}_{t+1,e_1}$, $\hat{\textbf{z}}_{t+1,e_2}, \dots, \hat{\textbf{z}}_{t+1,e_{|\mathcal{E}|}})^T$, we employ the decoder of an existing TKG extrapolation model (e.g., RE-GCN, TiRGN, and RETIA) to perform prediction at timestamp $t+1$. We formulate the scoring function measuring the likelihood of a candidate entity $e \in \mathcal{E}$ being the correct answer for a given query $(s, r_I, ?, t+1)$ as follows:
\begin{equation}
\phi(s,r_I,e,t+1)=\sigma_2(\hat{{\mathbf{z}}}_{t+1,e} \cdot \text{Decoder}(\hat{\mathbf{z}}_{t+1,s},\mathbf{r}_{t+1,r_I})),
\end{equation} 
where $\text{Decoder}$ denotes the decoder of the TKG extrapolation model and $\sigma_2$ is the Sigmoid function. $\hat{\mathbf{z}}_{t+1,s}$ is the embedding of the subject entity $s$ at timestamp $t+1$, $\mathbf{r}_{t+1,r_I}$ is the embedding of the relation $r_I$, and $\hat{\mathbf{z}}_{t+1,e}$ represents the embeddings of $e$. 
The loss $\mathcal{L}_{\text{tkg}}$ is formalized as follows:
\begin{equation}
\mathcal{L}_{\text{tkg}} = -\sum_{t=0}^{|T|-1}\sum_{(s,r_I,o,t+1)\in \mathcal{F}_I^{t+1}}\sum_{e\in\mathcal{E}}y_{t+1}^o\log{\phi(s,r_I,e,t+1)},
\end{equation}
where $y_{t+1}^o \in \mathbb{R}^{|\mathcal{E}|}$ is the label vector.
The final loss function is $\mathcal{L}=\mathcal{L}_{\text{tkg}}+\alpha_1\mathcal{L}_{\text{hie}}+\alpha_2\mathcal{L}_{\text{cl}}$,
where $\alpha_1$ and $\alpha_2$ are hyperparameters that balance the contributions of the different loss components.

\begin{table*}[t]
\centering
\caption{Performance of models on the task of TKG extrapolation. The best results are boldfaced.}
\vspace{-4mm}
\renewcommand{\arraystretch}{0.3}
\setlength{\tabcolsep}{4.8pt}
\begin{tikzpicture}
\node[
    inner sep=1.5pt,
    outer sep=1.5pt
] (tableimage) {
\resizebox{0.93\textwidth}{!}{
\begin{tabular}{cl|ccc|ccc|ccc|ccc|ccc}
\hline
\multirow{2}{*}{\emph{\textbf{Category}}} &\multirow{2}{*}{\emph{\textbf{Method}}} &\multicolumn{3}{c|}{\cellcolor[HTML]{F2F2F2}\emph{\textbf{ICEWS14}}} & \multicolumn{3}{c|}{\cellcolor[HTML]{F2F2F2}\emph{\textbf{ICEWS18}}} & \multicolumn{3}{c|}{\cellcolor[HTML]{F2F2F2}\emph{\textbf{ICEWS05-15}}} & \multicolumn{3}{c|}{\cellcolor[HTML]{F2F2F2}\emph{\textbf{GDELT}}} & \multicolumn{3}{c}{\cellcolor[HTML]{F2F2F2}\emph{\textbf{Wiki35k}}} \\
\cline{3-17}
& & \emph{MRR} & \emph{H@1} & \emph{H@10} & \emph{MRR} & \emph{H@1} & \emph{H@10} & \emph{MRR} & \emph{H@1} & \emph{H@10} & \emph{MRR} & \emph{H@1} & \emph{H@10} & \emph{MRR} & \emph{H@1} & \emph{H@10} \\
\hline

base model & RE-GCN~\cite{regcn} (SIGIR'21) &.422 &.319 &.620 &.325 &.223 &.522 &.480 &.373 &.685 &.195 &.122 &.335 & .576 & .493 & .750 \\
baseline & \quad - JOIE~\cite{joie} (KDD'19)&.423 &.322 &.620 &.326 &.220 &.531 &.484 &.374 &.693 &.195 &.124 &.334 & .577 & .493 & .750 \\
baseline & \quad - HyperCL~\cite{hypercl} (ACL'24)&.420 &.317 &.624 &.323 &.219 &.528 &.485 &.375 &.696 &.194 &.123 &.333 & .579 & .495 & .751 \\
baseline & \quad - CISS~\cite{ciss} (KDD'24) & .425 & .322 & .624 & .330 & .225 & .536 & .502 & .385 & .712 & .196 & .123 & .335 & .584 & .504 & .752 \\
baseline & \quad - LLM-DA~\cite{llmda} (NeurIPS'24) &.461 &.356 &.662 &.320 &.218 &.519 &.501 &.394 &.710 & .178 & .104 & .318 & .560 & .478 & .736 \\
baseline & \quad - ANEL~\cite{anel} (WWW'25) &.432 &.329 &.631 &.335 &.231 &.531 &.492 &.384 &.698 &.203 &.128 &.350 & .588 & .501 & .760 \\
\rowcolor[HTML]{E6E6E6} ours& \quad - \textit{OntoTKGE}&\textbf{.541} &\textbf{.415} &\textbf{.782} &\textbf{.371} &\textbf{.250} &\textbf{.615} &\textbf{.609} &\textbf{.487} &\textbf{.840} &\textbf{.206} &\textbf{.129} &\textbf{.356}  & \textbf{.689} & \textbf{.623} & \textbf{.786} \\
\hline

base model & TiRGN~\cite{tirgn} (IJCAI'22) &.446 &.343 &.647 &.333 &.229 &.540 &.500 &.391 &.711 &.215 &.136 &.374 & .597 & .516 & .768 \\
baseline & \quad - JOIE &.452 &.343 &.664 &.329 &.225 &.533 &.504 &.392 &.717 &.215 &.134 &.374 & .590 & .508 & .764 \\
baseline & \quad - HyperCL &.452 &.343 &.663 &.327 &.223 &.529 &.506 &.394 &.720 &.214 &.134 &.374 & .591 & .507 & .766 \\
baseline & \quad - CISS & .455 & .348 & .670 & .332 & .229 & .548 & .516 & .408 & .732 & .218 & .138 & .380 & .588 & .505 & .762 \\
baseline & \quad - LLM-DA &.471 &.369 &.671 &.357 &.255 &.570 &.521 &.416 &.728 &.207 &.129 &.366 & .590 & .509 & .763 \\
baseline & \quad - ANEL  &.447 &.344 &.646 &.340 &.235 &.547 &.505 &.396 &.712 &.220 &.138 &.383 & .600 & .518 & .782 \\
\rowcolor[HTML]{E6E6E6} ours& \quad - \textit{OntoTKGE} &\textbf{.562}&\textbf{.438} &\textbf{.798} &\textbf{.390} &\textbf{.268} &\textbf{.634} &\textbf{.624} &\textbf{.502} &\textbf{.853} &\textbf{.231} &\textbf{.142} &\textbf{.405}  & \textbf{.689} & \textbf{.620} & \textbf{.804} \\
\hline

base model & RETIA~\cite{retia} (ICDE'23) &.422 &.320 &.624 &.323 &.220 &.528 &.472 &.366 &.676 &.199 &.124 &.341 & .588 & .511 & .760 \\
baseline & \quad - JOIE &.423 &.317 &.630 &.319 &.217 &.519 &.472 &.366 &.676 &.196 &.124 &.337 & .590 & .512 & .761 \\
baseline & \quad - HyperCL &.427 &.326 &.627 &.320 &.218 &.521 &.471 &.366 &.675 &.196 &.124 &.336 & .577 & .498 & .751 \\
baseline & \quad - CISS & .423 & .318 & .628 & .322 & .216 & .533 & .477 & .370 & .683 & .199 & .125 & .341 & .591 & .513 & .758 \\
baseline & \quad - LLM-DA &.408 &.312 &.602 &.243 &.151 &.430 &.433 &.333 &.632 &.181 &.105 &.321 & .570 & .492 & .742 \\
baseline & \quad - ANEL & .420 & .318 & .623 & .334 & .234 & .540 & .488 & .380 & .691 & .201 & .124 & .348 & .601 & .524 & .773 \\
\rowcolor[HTML]{E6E6E6} ours& \quad - \textit{OntoTKGE} &\textbf{.554} &\textbf{.426} &\textbf{.797} &\textbf{.384} &\textbf{.264} &\textbf{.623} &\textbf{.616} &\textbf{.494} &\textbf{.846} &\textbf{.205} &\textbf{.127} &\textbf{.355} & \textbf{.695} & \textbf{.625} & \textbf{.810} \\
\hline

base model & LogCL~\cite{logcl} (ICDE'24) &.485 &.368 &.715 &.351 &.243 &.574 &.569 &.458 &.742 &.237 &.146 &.423 & .745 & .705 & .812 \\
baseline & \quad - JOIE &.488 &.372 &.714 &.355 &.241 &.581 &.562 &.452 &.742 &.236 &.145 &.422 & .742 & .702 & .813 \\
baseline & \quad - HyperCL &.488 &.371 &.715 &.356 &.243 &.581 &.561 &.450 &.742 &.235 &.145 &.420 & .755 & .719 & .821 \\
baseline & \quad - CISS & .484 & .372 & .708 & .360 & .246 & .582 & .572 & .460 & .745 & .238 & .146 & .426 & .752 & .744 & .822 \\
baseline & \quad - LLM-DA &.496 &.377 &.719 &.351 &.241 &.565 &.568 &.458 &.744 &.230 &.139 &.414 & .738 & .697 & .807 \\
baseline & \quad - ANEL & .490 & .372 & .718 & .358 & .249 & .577 & .563 & .452 & .740 & .242 & \textbf{.147} & .430 & .750 & .709 & .816 \\
\rowcolor[HTML]{E6E6E6} ours& \quad - \textit{OntoTKGE}  &\textbf{.597} &\textbf{.472} &\textbf{.839} &\textbf{.426} &\textbf{.298} &\textbf{.681} &\textbf{.666} &\textbf{.543} &\textbf{.899} &\textbf{.244} &\textbf{.147} &\textbf{.438} & \textbf{.787} & \textbf{.758} & \textbf{.835} \\
\hline

base model & HisRES~\cite{hisres} (ICDE'25) &.495 &.385 &.693 &.363 &.250 &.586 &.587 &.483 &.779 &.256 &.161 &.452 & .534 & .461 & .683 \\
baseline & \quad - JOIE &.494 &.382 &.690 &.362 &.248 &.588 &.582 &.477 &.777 &.253 &.159 &.441 & .552 & .485 & .731 \\
baseline & \quad - HyperCL &.492 &.381 &.688 &.360 &.246 &.583 &.586 &.481 &.779 &.255 &.160 &.444 & .544 & .473 & .702 \\
baseline & \quad - CISS & .498 & .386 & .698 & .360 & .249 & .584 & .588 & .483 & .783 & .250 & .157 & .450 & .550 & .481 & .722 \\
baseline & \quad - LLM-DA &.500 &.396 &.694 &.369 &.259 &.581 &.591 &.487 &.779 &.249 &.154 &.443 & .528 & .454 & .677 \\
baseline & \quad - ANEL & .502 & .392 & .698 & .368 & .255 & .588 & .601 & .495 & .793 & .256 & .162 & .453 & .533 & .460 & .683 \\
\rowcolor[HTML]{E6E6E6} ours& \quad - \textit{OntoTKGE}
&\textbf{.601} &\textbf{.476} &\textbf{.838} &\textbf{.412} &\textbf{.284} &\textbf{.666} &\textbf{.700} &\textbf{.588} &\textbf{.906} &\textbf{.262} &\textbf{.164} &\textbf{.463}  & \textbf{.647} & \textbf{.577} & \textbf{.767} \\
\hline
\end{tabular}
}
};
\end{tikzpicture}
\vspace{-3mm}
\label{tab:TKGR_results}
\end{table*}

\vspace{-1mm}
\subsection{Complexity Analysis}
The global ontology-aware evolutionary encoder has a time complexity of $O(2|\mathcal{F}_O|)$, while the local ontology-aware relevance encoder has a time complexity of $O(|\hat{\mathcal{F}}_O|)$. The time complexity of contrastive learning is $O(|\mathcal{M}_{t+1}|^2)$. 
Thus, the time complexity of OntoTKGE is $O(2|\mathcal{F}_O| + |\hat{\mathcal{F}}_O| + |\mathcal{M}_{t+1}|^2)$. 
As $|\mathcal{F}_O|$, $|\hat{\mathcal{F}}_O|$, and $|\mathcal{M}_{t+1}|$ are mostly small in TKGs, OntoTKGE introduces only minimal additional complexity over its corresponding base models.
The time complexity of the ontology construction pipeline is $O(|\mathcal{E}|c_{\mathrm{link}}+Uc_{\mathrm{type}})$, where $U$ denotes the number of unlinked entities, with $c_{\mathrm{link}}$ and $c_{\mathrm{type}}$ denoting the maximum external calls per entity in the linking and typing stages. Since $U$ is typically small in practice,  
e.g., ICEWS14 contains 461 (6.5\%) unlinked entities, the overall time complexity scales approximately linearly with $|\mathcal{E}|$.

\vspace{-1mm}
\section{Experimental Study}
\subsection{Experiment Settings}
\noindent\textbf{Data Sets.} We conduct experiments over $5$ common benchmark data sets: ICEWS14 \cite{icews}, ICEWS18, ICEWS05-15, and GDELT \cite{gdelt}. 
To evaluate the impact of OntoTKGE, we construct Wiki35k from the Wikidata dump downloaded in July. Following \cite{wiki_benchmark}, we extract triples from the dump and retain point facts (i.e., facts annotated with point time) and interval facts (i.e., facts annotated with both start time and end time). 
We keep facts whose valid years overlap with 1829 to 2029 and truncate overlapping intervals to this range. We use one year as the time granularity. Point facts are assigned to their annotated years, while facts with complete validity intervals are expanded to all years covered by the intervals.
To cover diverse entity activity, we first compute three statistics for each entity: the number of facts involving the entity, the number of associated relation types, and the number of active years. Then, we sample a subset of popular entities with broad temporal and relational coverage and sparse entities with limited observations, retaining facts whose subject entity and object entity are both selected.
Wiki35k contains 34,808 entities, 199 relations, 201 timestamps, and 510,301 facts.
Following \cite{regcn}, each data set is divided into training, validation, and test sets with proportions of 80\%/10\%/10\%. 

\noindent\textbf{Evaluation Metrics.} Following previous TKG extrapolation studies \cite{regcn,tirgn,logcl,hisres}, we use the same metrics: mean reciprocal rank (MRR) and Hits@k (k=$1, 10$) abbreviated as H@k. 
We report the experimental results under the time-aware filtered setting as \cite{logcl,hisres}.


\noindent\textbf{Implementation Details.} We use official implementations of $5$ common TKG extrapolation models as OntoTKGE's base models, i.e., RE-GCN~\cite{regcn}, TiRGN~\cite{tirgn}, RETIA~\cite{retia}, LogCL~\cite{logcl}, and HisRES~\cite{hisres}.
OntoTKGE enhances the evolutionary encoders of RE-GCN and RETIA, as well as the local recurrent and global historical encoders of TiRGN, LogCL, and HisRES. For decoders, OntoTKGE uses Time-ConvTransE for TiRGN, while using ConvTransE for other models.
All models are trained for $30$ epochs. 
The random seed is set to $42$. 
We set the embedding dimension to $200$ and unify the data pre-processing following RE-GCN. 
We use GPT-4o-mini \cite{gpt4},  Qwen3-Embedding-0.6B \cite{qwen3-embedding}, and Wikidata \cite{wikidata} as the LLM, retrieval model, and static KG, respectively. $\alpha_1$, $\alpha_2$ and $\tau$ are set to $0.1$, $0.1$ and $0.07$ for all data sets, respectively. $K$, $N$ and $J$ are set to the best-performing values in Section \ref{subsec:param}. We use \textit{Sub} composition operation. 

\vspace{-3.5mm}
\subsection{Effectiveness Study}
Since there are no existing ontology-enhanced methods for TKG extrapolation, 
we select two categories of baselines: (1) TKG extrapolation enhancement methods ANEL~\cite{anel} and LLM-DA~\cite{llmda}; (2) ontology-enhanced static KG completion methods JOIE~\cite{joie},  HyperCL~\cite{hypercl}, and CISS~\cite{ciss} (which are adapted to the same temporal settings for fair comparison).
We reproduce LLM‑DA, JOIE, and HyperCL using the open‑source code. 
For CISS (no public code) and ANEL (partial code), we implement the former and complete the latter based on the corresponding paper descriptions.
From the results shown in Table \ref{tab:TKGR_results}, we observe that: (1) OntoTKGE consistently improves all base models (i.e., RE-GCN, TiRGN, RETIA, LogCL, and HisRES) used by OntoTKGE on all data sets; (2) OntoTKGE equipped with different TKG extrapolation models outperforms all baselines (i.e., JOIE, HyperCL, LLM-DA, ANEL, and CISS) on various data sets. These results demonstrate that OntoTKGE is flexible and can alleviate the sparse entity issue by leveraging ontological knowledge from the ontology-view KG. In addition, ontology-enhanced static KG completion methods tend to reduce performance, which may be attributed to the fact that (1) they do not incorporate ontological knowledge from the ontology-view KG to guide the model's learning process; (2) using a uniform ontological constraint to entities may hinder the model from capturing their evolutionary patterns.
The gains vary across data sets. It achieves larger improvements on ICEWS14, ICEWS18, ICEWS05-15, and Wiki35k, where sparse entities have fewer facts and the ontology-view KG enriches entity embeddings with ontological knowledge. On GDELT, lower entity sparsity makes historical facts more informative, reducing the relative contribution of the ontology-view KG.
Additionally, it is worth noting that for any given data set, the improvement of OntoTKGE over different base models is roughly the same, indicating that OntoTKGE is not very sensitive to the base model performance.

\begin{table}[t]
\renewcommand{\arraystretch}{0.55}
\centering
\caption{Results of ablation study.}
\vspace{-4mm}
\setlength{\tabcolsep}{1.2pt}
\begin{tikzpicture}
\node[
    inner sep=1.5pt,
    outer sep=1.5pt
] (tableimage) {
\scalebox{0.66}{
\begin{tabular}{l|ccc|ccc|ccc|ccc}
\hline
\multirow{2}{*}{\emph{\textbf{Method}}} & \multicolumn{3}{c|}{\cellcolor[HTML]{F2F2F2}\emph{\textbf{ICEWS14}}} & \multicolumn{3}{c|}{\cellcolor[HTML]{F2F2F2}\emph{\textbf{ICEWS18}}} & \multicolumn{3}{c|}{\cellcolor[HTML]{F2F2F2}\emph{\textbf{ICEWS05-15}}} & \multicolumn{3}{c}{\cellcolor[HTML]{F2F2F2}\emph{\textbf{Wiki35k}}} \\
\cline{2-13}
 & \emph{MRR} & \emph{H@1} & \emph{H@10} & \emph{MRR} & \emph{H@1} & \emph{H@10} & \emph{MRR} & \emph{H@1} & \emph{H@10} & \emph{MRR} & \emph{H@1} & \emph{H@10} \\
\hline

\rowcolor[HTML]{E6E6E6} RE-GCN-\textit{OntoTKGE} & \textbf{.541} & \textbf{.415} & \textbf{.782} & \textbf{.371} & \textbf{.250} & \textbf{.615} & \textbf{.609} & \textbf{.487} & \textbf{.840}  & \textbf{.689} & .623 & \textbf{.786} \\
\quad w/o $\mathcal{L}_{cl}$ &.528 &.402 &.774 & .358 & .242 & .593 & .592 & .469 & .827  & .687 & \textbf{.625} & .782 \\
\quad w/o $\mathcal{L}_{hie}$ &.533 &.405 &.785 & .361 & .243 & .597 & .599 & .472 & .837  & .674 & .606 & .786 \\
\quad w/o L-Enc &.426 &.322 &.628 & .327 & .225 & .527 & .478 & .369 & .690  & .576 & .493 & .746 \\
\quad w/o G-Enc & .498 & .374 & .741 & .335 & .223 & .561 & .537 & .410 & .785  & .047 & .024 & .084 \\

\hline

\rowcolor[HTML]{E6E6E6} LogCL-\textit{OntoTKGE} & \textbf{.597} & \textbf{.472} & \textbf{.839} & \textbf{.426} & \textbf{.298} & \textbf{.681} & \textbf{.666} & \textbf{.543} & \textbf{.899}  & \textbf{.787} & \textbf{.758} & \textbf{.835} \\
\quad w/o $\mathcal{L}_{cl}$ & .584 & .457 & .832 & .388 & .263 & .636 & .648 & .521 & .891  & .421 & .336 & .582 \\
\quad w/o $\mathcal{L}_{hie}$ & .590 & .463 & .836 & .414 & .289 & .670 & .662 & .540 & .897  & .779 & .756 & .834 \\
\quad w/o L-Enc & .489 & .370 & .712 & .354 & .237 & .583 & .551 & .432 & .777  & .773 & .741 & .827 \\
\quad w/o G-Enc & .498 & .374 & .741 & .335 & .223 & .561 & .537 & .410 & .785  & .047 & .024 & .084 \\
\hline

\rowcolor[HTML]{E6E6E6} HisRES-\textit{OntoTKGE} & \textbf{.601} & \textbf{.476} & \textbf{.838} & \textbf{.412} & \textbf{.284} & \textbf{.666} & \textbf{.700} & \textbf{.588} & \textbf{.906}  & \textbf{.647} & \textbf{.577} & \textbf{.767} \\
\quad w/o $\mathcal{L}_{cl}$ &.572 &.455 &.812 &.390 &.265 &.642 &.656 &.532 &.889  & .599 & .521 & .731 \\
\quad w/o $\mathcal{L}_{hie}$&.591 &.467 &.816 &.408 &.281 &.661 &.688 &.576 &.878  & .641 & .570 & .760 \\
\quad w/o L-Enc &.496 &.389 &.701 &.367 &.255 &.594 &.589 &.486 &.782  & .536 & .460 & .694 \\
\quad w/o G-Enc & .498 & .374 & .741 & .335 & .223 & .561 & .537 & .410 & .785  & .047 & .024 & .084 \\

\hline
\end{tabular}
}
};
\end{tikzpicture}
\vspace{-6mm}
\label{tab:ablation}
\end{table}

\begin{table*}[t]
\centering
\caption{TKG extrapolation performance across different entity degree ranges over ICEWS14.}
\vspace{-4mm}
\setlength{\tabcolsep}{3pt}
\renewcommand{\arraystretch}{0.4}

\newcolumntype{C}{>{\centering\arraybackslash}p{0.72cm}}

\scalebox{0.82}{
\begin{tabular}{ll|*{3}{C}|*{3}{C}|*{3}{C}|*{3}{C}|*{3}{C}|*{3}{C}}
\hline
\multicolumn{2}{c}{\textbf{\emph{Method}}} & 
\multicolumn{3}{c|}{\cellcolor[HTML]{F2F2F2}\textbf{\emph{RE-GCN}}} & \multicolumn{3}{c|}{\cellcolor[HTML]{F2F2F2}\textbf{\emph{RE-GCN-OntoTKGE}}} & 
\multicolumn{3}{c|}{\cellcolor[HTML]{F2F2F2}\textbf{\emph{LogCL}}} & \multicolumn{3}{c|}{\cellcolor[HTML]{F2F2F2}\textbf{\emph{LogCL-OntoTKGE}}} & 
\multicolumn{3}{c|}{\cellcolor[HTML]{F2F2F2}\textbf{\emph{HisRES}}} & \multicolumn{3}{c}{\cellcolor[HTML]{F2F2F2}\textbf{\emph{HisRES-OntoTKGE}}} \\

\cmidrule(lr){1-2} \cmidrule(lr){3-5} \cmidrule(lr){6-8} \cmidrule(lr){9-11} \cmidrule(lr){12-14} \cmidrule(lr){15-17} \cmidrule(lr){18-20}

\textbf{\emph{Degree}} & \textbf{\emph{Num}} & 
\emph{MRR} & \emph{H@1} & \emph{H@10} & \emph{MRR} & \emph{H@1} & \emph{H@10} & 
\emph{MRR} & \emph{H@1} & \emph{H@10} & \emph{MRR} & \emph{H@1} & \emph{H@10} & 
\emph{MRR} & \emph{H@1} & \emph{H@10} & \emph{MRR} & \emph{H@1} & \emph{H@10} \\
\midrule

$[0, 10]$ & 6,292 & 
.278 &.194 &.443 & \cellcolor[HTML]{E6E6E6}.564 & \cellcolor[HTML]{E6E6E6}.433 & \cellcolor[HTML]{E6E6E6}.807 & 
.380 &.274 &.585 & \cellcolor[HTML]{E6E6E6}.659 & \cellcolor[HTML]{E6E6E6}.534 & \cellcolor[HTML]{E6E6E6}.880 & 
.361 &.273 &.536 & \cellcolor[HTML]{E6E6E6}.644 & \cellcolor[HTML]{E6E6E6}.518 & \cellcolor[HTML]{E6E6E6}.886 \\

$[10, 20]$ & 350 & 
.356 &.249 &.568 & \cellcolor[HTML]{E6E6E6}.665 & \cellcolor[HTML]{E6E6E6}.539 & \cellcolor[HTML]{E6E6E6}.902 & 
.459 &.328 &.701 & \cellcolor[HTML]{E6E6E6}.762 & \cellcolor[HTML]{E6E6E6}.646 & \cellcolor[HTML]{E6E6E6}.956 & 
.454 &.336 &.671 & \cellcolor[HTML]{E6E6E6}.749 & \cellcolor[HTML]{E6E6E6}.629 & \cellcolor[HTML]{E6E6E6}.960 \\

$[20, 30]$ & 147 & 
.391 &.286 &.603 & \cellcolor[HTML]{E6E6E6}.704 & \cellcolor[HTML]{E6E6E6}.582 & \cellcolor[HTML]{E6E6E6}.927 & 
.522 &.395 &.742 & \cellcolor[HTML]{E6E6E6}.787 & \cellcolor[HTML]{E6E6E6}.684 & \cellcolor[HTML]{E6E6E6}.965 & 
.515 &.407 &.724 & \cellcolor[HTML]{E6E6E6}.781 & \cellcolor[HTML]{E6E6E6}.666 & \cellcolor[HTML]{E6E6E6}.977 \\

$[30, 40]$ & 80 & 
.361 &.266 &.536 & \cellcolor[HTML]{E6E6E6}.670 & \cellcolor[HTML]{E6E6E6}.538 & \cellcolor[HTML]{E6E6E6}.952 & 
.495 &.375 &.707 & \cellcolor[HTML]{E6E6E6}.747 & \cellcolor[HTML]{E6E6E6}.625 & \cellcolor[HTML]{E6E6E6}.972 & 
.475 &.370 &.664 & \cellcolor[HTML]{E6E6E6}.735 & \cellcolor[HTML]{E6E6E6}.610 & \cellcolor[HTML]{E6E6E6}.967 \\

$[40, 50]$ & 45 & 
.414 &.297 &.649 & \cellcolor[HTML]{E6E6E6}.751 & \cellcolor[HTML]{E6E6E6}.615 & \cellcolor[HTML]{E6E6E6}.975 & 
.562 &.437 &.799 & \cellcolor[HTML]{E6E6E6}.822 & \cellcolor[HTML]{E6E6E6}.716 & \cellcolor[HTML]{E6E6E6}.985 & 
.531 &.414 &.762 & \cellcolor[HTML]{E6E6E6}.809 & \cellcolor[HTML]{E6E6E6}.696 & \cellcolor[HTML]{E6E6E6}.985 \\

$[50, 100]$ & 93 & 
.466 &.365 &.655 & \cellcolor[HTML]{E6E6E6}.705 & \cellcolor[HTML]{E6E6E6}.585 & \cellcolor[HTML]{E6E6E6}.928 & 
.597 &.483 &.797 & \cellcolor[HTML]{E6E6E6}.774 & \cellcolor[HTML]{E6E6E6}.661 & \cellcolor[HTML]{E6E6E6}.958 & 
.579 &.474 &.767 & \cellcolor[HTML]{E6E6E6}.771 & \cellcolor[HTML]{E6E6E6}.660 & \cellcolor[HTML]{E6E6E6}.960 \\

$[100, \text{max}]$ & 121 & 
.479 &.374 &.678 & \cellcolor[HTML]{E6E6E6}.575 & \cellcolor[HTML]{E6E6E6}.453 & \cellcolor[HTML]{E6E6E6}.814 & 
.548 &.430 &.779 & \cellcolor[HTML]{E6E6E6}.626 & \cellcolor[HTML]{E6E6E6}.501 & \cellcolor[HTML]{E6E6E6}.875 & 
.559 &.450 &.762 & \cellcolor[HTML]{E6E6E6}.630 & \cellcolor[HTML]{E6E6E6}.508 & \cellcolor[HTML]{E6E6E6}.862 \\
\hline
\end{tabular}} 
\vspace{-3.5mm}
\label{tab:degree}
\end{table*}





\begin{figure*}[t]
    \centering
    \begin{tikzpicture}
        \node[
            inner sep=0pt,
            outer sep=0pt
        ] (image) {
            \includegraphics[
                width=0.9\linewidth
            ]{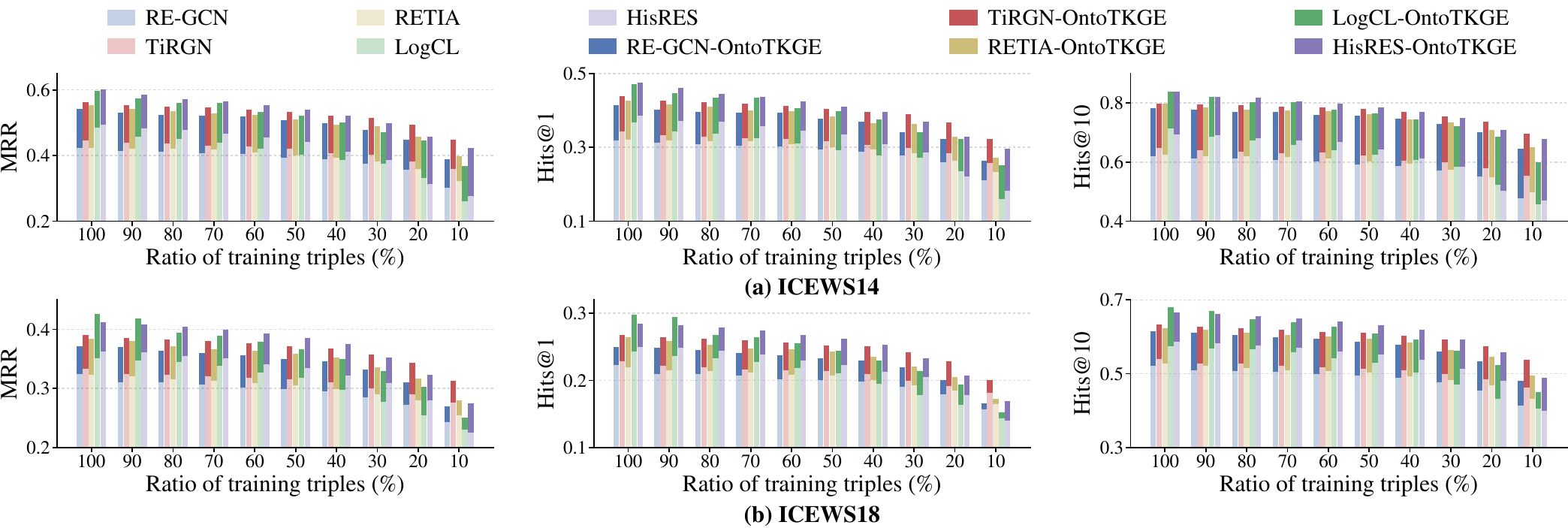}
        };
    \end{tikzpicture}
    \vspace{-4mm}
    \caption{Scalability evaluation over ICEWS14 and ICEWS18.}
    \vspace{-5mm}
    \label{fig:sparse}
\end{figure*}

\begin{table}[t]
\renewcommand{\arraystretch}{0.6}
\setlength{\tabcolsep}{1.2pt}
\centering
\caption{Performance of OntoTKGE's different variants.}
\vspace{-4mm}
\begin{tikzpicture}
\node[
    inner sep=1.5pt,
    outer sep=1.5pt
] (tableimage) {
\scalebox{0.63}{
\begin{tabular}{l|ccc|ccc|ccc|ccc}
\hline
\multirow{2}{*}{\emph{\textbf{Variants}}} & \multicolumn{3}{c|}{\cellcolor[HTML]{F2F2F2}\emph{\textbf{ICEWS14}}} & \multicolumn{3}{c|}{\cellcolor[HTML]{F2F2F2}\emph{\textbf{ICEWS18}}} & \multicolumn{3}{c|}{\cellcolor[HTML]{F2F2F2}\emph{\textbf{ICEWS05-15}}} & \multicolumn{3}{c}{\cellcolor[HTML]{F2F2F2}\emph{\textbf{Wiki35k}}} \\
\cline{2-13}
& \emph{MRR} & \emph{H@1} & \emph{H@10} & \emph{MRR} & \emph{H@1} & \emph{H@10} & \emph{MRR} & \emph{H@1} & \emph{H@10} & \emph{MRR} & \emph{H@1} & \emph{H@10} \\
\hline

\rowcolor[HTML]{E6E6E6} RE-GCN-\textit{OntoTKGE} & \textbf{.541} & \textbf{.415} & \textbf{.782} & \textbf{.371} & \textbf{.250} & \textbf{.615} & \textbf{.609} & \textbf{.487} & \textbf{.840}  & \textbf{.689} & \textbf{.623} & \textbf{.786} \\
\quad RE-GCN-\textit{OntoTKGE}$_{CompGCN}$ &.532 &.404 &.783 &.368 &.248 & .610 &.600 &.484 &.831  & .680 & .608 & .785 \\
\quad RE-GCN-\textit{OntoTKGE}$_{RGCN}$ &.527 &.405 &.763 &.369 &.250 &.602 &.576 &.459 &.797  & .662 & .590 & .784 \\
\quad RE-GCN-\textit{OntoTKGE}$_{KRAT}$ &.531 &.403 &.776 &.370 &.251 &.611 &.592 &.470 &.814  & .681 & .615 & .789 \\
\quad RE-GCN-\textit{OntoTKGE}$_{RandInit}$ &.529 &.402 &.773 &.357 &.239 &.596 &.594 &.472 &.827  & .676 & .609 & .783 \\
\quad RE-GCN-\textit{OntoTKGE}$_{Sum}$ &.527 &.400 &.775 & .360 & .242 & .597 & .587 & .463 & .826  & .670 & .606 & .774 \\
\quad RE-GCN-\textit{OntoTKGE}$_{Mult}$ &.523 &.398 &.762 & .368 & .251 & .601 & .580 & .463 & .800  & .671 & .603 & .786 \\
\quad RE-GCN-\textit{OntoTKGE}$_{Corr}$ &.524 &.401 &.761 & .366 & .249 & .602 & .572 & .454 & .796  & .673 & .604 & .783 \\

\hline

\rowcolor[HTML]{E6E6E6} LogCL-\textit{OntoTKGE} &\textbf{.597} &\textbf{.472} &\textbf{.839} &\textbf{.426} & \textbf{.298} &\textbf{.681} &\textbf{.666} &\textbf{.543} & \textbf{.899}  & \textbf{.787} & .758 & \textbf{.835} \\
\quad LogCL-\textit{OntoTKGE}$_{CompGCN}$ & .590 & .467 & .831 & .422 & .295 & .678 & .651 & .533 & .874  & .785 & .756 & .830 \\
\quad LogCL-\textit{OntoTKGE}$_{RGCN}$ & .583 & .459 & .822 & .409 & .283 & .660 & .631 & .510 & .860  & .785 & .756 & .834 \\
\quad LogCL-\textit{OntoTKGE}$_{KRAT}$ & .577 & .452 & .817 & .390 & .266 & .638 & .634 & .509 & .871  & .779 & .747 & .833 \\
\quad LogCL-\textit{OntoTKGE}$_{RandInit}$ & .579 & .452 & .828 & .380 & .256 & .625 & .643 & .516 & .885  & .764 & .729 & .826 \\
\quad LogCL-\textit{OntoTKGE}$_{Sum}$ & .578 & .451 & .824 & .402 & .275 & .658 & .646 & .517 & .889  & .779 & .749 & .828 \\
\quad LogCL-\textit{OntoTKGE}$_{Mult}$ &.585 &.459 &.826 &.408 &.282 &.660 &.632 &.510 &.865  & .786 & \textbf{.759} & .835 \\
\quad LogCL-\textit{OntoTKGE}$_{Corr}$ &.579 &.453&.818 &.396 &.272 &.643 &.628 &.507 &.862  & .785 & .756 & .833 \\

\hline

\rowcolor[HTML]{E6E6E6} HisRES-\textit{OntoTKGE} & .601 & .476 & \textbf{.838} & .412 & .284 & .666 & \textbf{.700} & \textbf{.588} & \textbf{.906}  & \textbf{.647} & .577 & \textbf{.767} \\
\quad HisRES-\textit{OntoTKGE}$_{CompGCN}$ &.596 &.470 &.822 &.404 &.276 &.658 &.695 &.582 &.892  & .638 & .570 & .761 \\
\quad HisRES-\textit{OntoTKGE}$_{RGCN}$ &.580 &.458 &.808 &.402 &.278 &.648 &.665 &.557 &.866  & .641 & .571 & .764 \\
\quad HisRES-\textit{OntoTKGE}$_{KRAT}$ &.575 &.452 &.803 &.399 &.275 &.646 &.671 &.558 &.882  & .641 & .571 & .762 \\
\quad HisRES-\textit{OntoTKGE}$_{RandInit}$ &.580 &.456 &.811 &.403 &.278 &.655 &.686 &.573 &.896  & .632 & .567 & .755 \\
\quad HisRES-\textit{OntoTKGE}$_{Sum}$ &.579 &.449 &.820 &.399 &.273 &.650 &.676 &.560 &.894  & .639 & .568 & .763 \\
\quad HisRES-\textit{OntoTKGE}$_{Mult}$ &\textbf{.603} &\textbf{.479} &.836 &\textbf{.415} &\textbf{.289} &\textbf{.668} &.694 &.581 &.905  & .642 & .575 & .763 \\
\quad HisRES-\textit{OntoTKGE}$_{Corr}$ &.589 &.464 &.823 &.402 &.277 &.650 &.682 &.566 &.897  & .646 & \textbf{.580} & .764 \\

\hline
\end{tabular}
}
};
\end{tikzpicture}
\vspace{-12mm}
\label{tab:ablation_variants}
\end{table}

\vspace{-2mm}
\subsection{Effect Analysis of Different Components} \label{subsec:each_component}
\textbf{Ablation Study. }\label{subsubsec:ablation_study}
To validate the importance of OntoTKGE's different components, we conduct ablation studies by removing G-Enc (i.e., the global ontology-aware evolutionary encoder), L-Enc (i.e., the local ontology-aware relevance encoder), hierarchical entailment loss $\mathcal{L}_{\text{hie}}$, and contrastive loss $\mathcal{L}_{\text{cl}}$. Table \ref{tab:ablation} shows that removing any component usually decreases performance, with varying impacts. 
For RE-GCN on Wiki35k, $\mathcal{L}_{\text{cl}}$ leads to a slight decrease in Hits@1, possibly because contrastive loss amplifies the noise from the less effective L-Enc.
The degradation caused by removing L-Enc confirms the value of modeling ontology neighborhoods associated with queries, which provide additional evidence for sparse entity prediction. The stronger impact of removing G-Enc on Wiki35k may be attributed to its yearly time granularity, where each snapshot involves more active entities and makes OntoTKGE without G-Enc more susceptible to irrelevant concepts. The result indicates the importance of G-Enc for ontology-view KGs with broader coverage and a longer temporal range. Removing $\mathcal{L}_{\text{hie}}$ or $\mathcal{L}_{\text{cl}}$ often causes smaller yet consistent drops. The former preserves the hierarchical structure among concepts, and the latter regularizes the representations learned from different ontology-aware encoders.

\noindent\textbf{Comparison with Different Variants. }\label{subsubsec:variants}
We define 7 different variants of OntoTKGE shown in Table \ref{tab:ablation_variants} by modifying key modules: (1) replacing our modified \textit{CompGCN} with traditional \textit{CompGCN}, relational graph convolutional network (\textit{RGCN})~\cite{rgcn}, or knowledge relational attention network (\textit{KRAT})~\cite{krat}; (2) replacing ontology-guided initialization with random initialization (i.e., \textit{RandInit}); (3) replacing the gated unit with summation (i.e., \textit{Sum}); (4) replacing \textit{Sub} with \textit{Mult} or \textit{Corr} introduced in Section \ref{subsec:global encoder}. Table \ref{tab:ablation_variants} shows that the full model is the most stable in most cases. Replacing the modified \textit{CompGCN} with \textit{RGCN} or \textit{KRAT} leads to larger drops than with traditional \textit{CompGCN}, indicating the necessity of modeling diverse relational semantics across hierarchy levels. \textit{RandInit} performs worse than ontology-guided initialization, which confirms that the ontology prior provides more informative initial representations for temporal evolution. \textit{Sum} is consistently inferior to the gated unit, demonstrating the benefit of adaptively fusing G-Enc and L-Enc. For composition operations, \textit{Sub} achieves the best overall results, with \textit{Mult} remaining competitive on HisRES, indicating that translational composition is generally more suitable while the optimal operation may depend on the base model.



\begin{figure}[!t]
    \centering
    \begin{tikzpicture}
        \node[
            inner sep=0pt,
            outer sep=0pt
        ] (image) {
            \includegraphics[
                width=0.9\linewidth
            ]{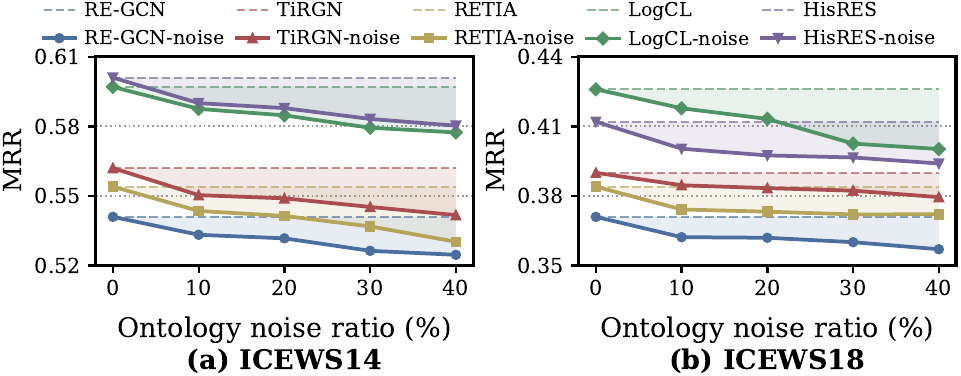}
        };
    \end{tikzpicture}
    \vspace{-4mm}
    \caption{Robustness evaluation.}
    \vspace{-7mm}
    \label{fig:ontology_noise}
\end{figure}

\begin{figure*}[t]
    \centering
    \begin{tikzpicture}
        \node[
            inner sep=0pt,
            outer sep=0pt
        ] (image) {
            \includegraphics[
                width=0.85\linewidth
            ]{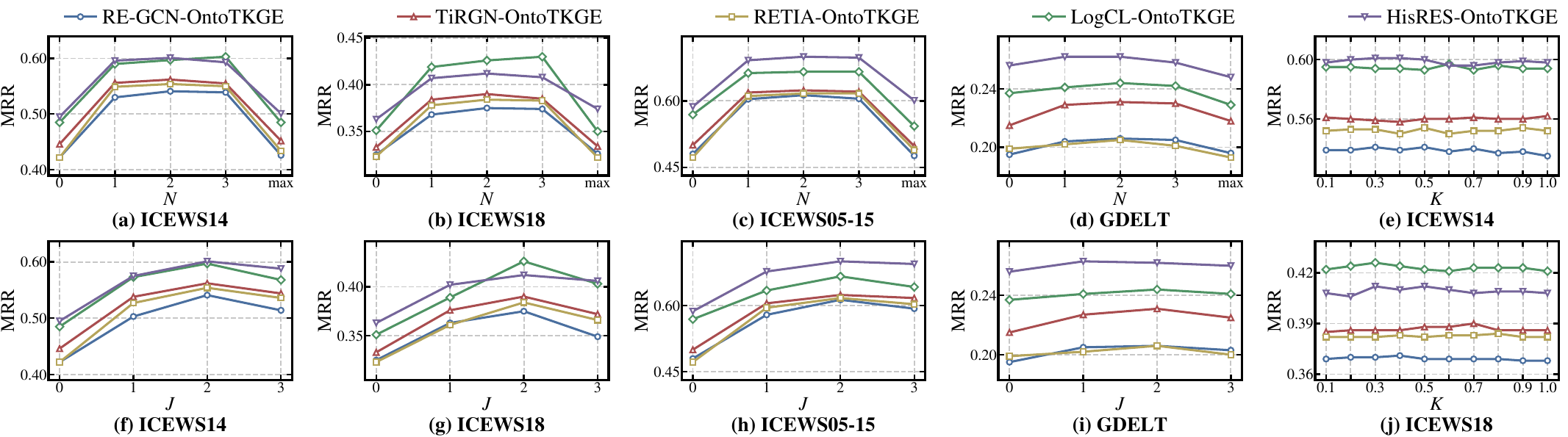}
        };
    \end{tikzpicture}
    \vspace{-4mm}
    \caption{Parameter sensitivity of OntoTKGE with different hop numbers $N$, numbers of layers $J$, and the parameter $K$.}
    \vspace{-5mm}
    \label{fig:param_combined}
\end{figure*}

\vspace{-2mm}
\subsection{Robustness Study}
\noindent\textbf{Entity Degree Analysis.}
Following \cite{krat}, we analyze the performance of TKG extrapolation across entities with different entity degrees over ICEWS14. 
Results in Table \ref{tab:degree} show OntoTKGE using different base models (i.e., RE-GCN-OntoTKGE, LogCL-OntoTKGE, and HisRES-OntoTKGE) yield significant improvements across all entity degree ranges compared with the corresponding base models (i.e., RE-GCN, LogCL, and HisRES). It is worth mentioning that the improvement is most pronounced for sparse entities, since the ontology-view KG allows them to effectively learn from related popular entities, alleviating the issue of data scarcity.

\noindent\textbf{Scalability Analysis.}
In this subsection, we evaluate the scalability of OntoTKGE by training it over different subsets of ICEWS14's training set and testing it over ICEWS14's test set. From the results shown in Figure \ref{fig:sparse}, we can see the performance of all models declines as the size of the training set decreases. However, OntoTKGE consistently improves all base models (i.e., RE-GCN, TiRGN, RETIA, LogCL, and HisRES), demonstrating OntoTKGE's robustness.

\noindent\textbf{Ontology Robustness Analysis.}
To test OntoTKGE's robustness under imperfect ontology construction, we inject noise into ICEWS14 and ICEWS18 with noise ratios ranging from 0\% to 40\% by replacing the object concepts in sampled ontological facts with incorrect ones.
Figure \ref{fig:ontology_noise} shows that increasing noise leads to a slight decline in performance. 
This demonstrates OntoTKGE is not highly sensitive to noisy ontology-view KGs,indicating that the global and local ontology-aware encoders preserve useful ontological information and maintain robust performance under imperfect ontology.

\begin{table}[t]
\centering
\caption{Construction quality and cost of ontology-view KGs.}
\label{tab:onto_construction_quality_cost}
\vspace{-4mm}
\renewcommand{\arraystretch}{0.8}
\setlength{\tabcolsep}{1.15pt}
\begingroup
\footnotesize
\begin{tikzpicture}
\node[
    inner sep=1.5pt,
    outer sep=1.5pt
] (tableimage) {
\resizebox{0.9\linewidth}{!}{
\begin{tabular}{@{}ll|cccccc|ccccc@{}}
\hline
\multirow{2}{*}{\emph{Setting}} & \multirow{2}{*}{\emph{Data set}}
& \multicolumn{6}{c|}{\cellcolor[HTML]{F2F2F2}\emph{Quality}} & \multicolumn{5}{c}{\cellcolor[HTML]{F2F2F2}\emph{Cost}} \\
\cline{3-13}
& & \emph{Link(\%)} & \emph{EL-Acc} & \emph{L-MiF1} & \emph{L-MaF1} & \emph{U-MiF1} & \emph{U-MaF1} & \emph{Ent} & \emph{Time(h)} & \emph{LLM} & \emph{API} & \emph{SPARQL} \\
\hline
\multirow{4}{*}{\shortstack{w/\\LLM}}
& ICEWS14    & 93.5 & .979 & .518 & .367 & .536 & .425 & 7,128  & 3.03 & 7,232  & 9,976  & 7,197 \\
& ICEWS18    & 94.6 & .976 & .505 & .332 & .670 & .438 & 23,033 & 6.10 & 17,894 & 24,562 & 14,386 \\
& ICEWS05-15 & 91.1 & .977 & .531 & .362 & .628 & .407 & 10,488 & 3.81 & 10,696 & 14,377 & 8,732 \\
& GDELT      & 86.7 & .966 & .468 & .378 & .253 & .372 & 7,691  & 1.15 & 2,890  & 3,970  & 4,295 \\
\hline
\multirow{4}{*}{\shortstack{w/o\\LLM}}
& ICEWS14    & 99.4 & .855 & .496 & .333 & .009 & .010 & 7,128  & 0.20 & 0 & 9,968  & 8,697 \\
& ICEWS18    & 99.6 & .864 & .481 & .298 & .004 & .005 & 23,033 & 0.61 & 0 & 24,656 & 17,041 \\
& ICEWS05-15 & 99.0 & .830 & .480 & .315 & .008 & .009 & 10,488 & 0.33 & 0 & 14,478 & 11,291 \\
& GDELT      & 99.9 & .831 & .453 & .301 & .000 & .000 & 7,691  & 0.15 & 0 & 4,024  & 5,540 \\
\hline
\end{tabular}}
};
\end{tikzpicture}
\endgroup
\vspace{-3mm}
\end{table}


\begin{figure}[!t]
    \centering
    \begin{tikzpicture}
        \node[
            inner sep=0pt,
            outer sep=0pt
        ] (image) {
            \includegraphics[
                width=0.9\linewidth
            ]{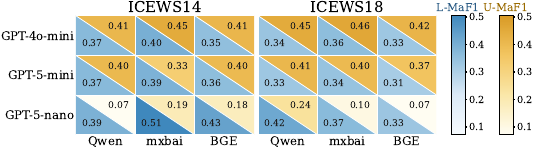}
        };
    \end{tikzpicture}
    \vspace{-4mm}
    \caption{Effect of different models for ontology construction}
    \vspace{-6mm}
    \label{fig:model_choice}
\end{figure}

\vspace{-2mm}
\subsection{Parameter Sensitivity}\label{subsec:param}
\noindent\textbf{Effect of the Hop Number $N$.}
We conduct parameter analysis to understand the impact of $N$. 
Results in Figure \ref{fig:param_combined} show that OntoTKGE equipped with different TKG extrapolation models achieves optimal results within a certain range (i.e., $N \in \{1, 2, 3\}$) but declines at the full ontology-view KG (denoted $\max$) over different data sets in terms of MRR. This indicates that a moderate subgraph provides necessary ontological facts related to the target entity, whereas an excessive one may introduce distant concepts.

\noindent\textbf{Effect of the Number of Layers $J$.}
We study the impact of $J$ on OntoTKGE's performance. 
Figure \ref{fig:param_combined} shows its performance with $J \in \{0, 1, 2, 3\}$ over 4 data sets. 
Performance at $J=0$ (ontological facts are used without graph propagation) is lower than in settings with propagation layers. $J=2$ achieves the best results in almost all data sets, while $J=3$ causes a slight drop. This indicates that 2 layers sufficiently capture the ontological knowledge of neighboring facts, while deeper propagation may introduce irrelevant concepts.

\noindent\textbf{Effect of the Parameter $K$.}
We evaluate OntoTKGE's sensitivity to $K$ (0.1-1.0) on ICEWS14/ICEWS18 (Figure~\ref{fig:param_combined}).
Performance remains stable across $K$ and base models. As the aperture of an entailment cone is related to its model capacity for hierarchical structures~\cite{hierarchical-entailment-learning-18}, this stability indicates that the considered range of $K$ suffices for modeling entailment relations in the ontology-view KG. This is supported by ablation results (Table~\ref{tab:ablation}), where removing $\mathcal{L}_{\text{hie}}$ causes the smallest drop in most settings, indicating that the low sensitivity to $K$ is consistent with the relatively small contribution of $\mathcal{L}_{\text{hie}}$.

\subsection{Ontology Construction Study}
%
\noindent\textbf{Ontology Construction Quality.}
We evaluate ontology-view KG construction using $Link$ (the percentage of linked entities), EL-Acc (entity linking accuracy~\cite{refined}), and common entity typing metrics~\cite{Entity_typing1,Entity_typing2} micro-F1 (MiF1)/macro-F1 (MaF1). Specifically, to handle linked and unlinked TKG entities for entity typing, we report scores: L-MiF1/L-MaF1 for linked entities (entities with manually verified links to Wikidata entries, which provide the reference ontological facts) and U-MiF1/U-MaF1 for unlinked entities (entities without accepted links to Wikidata entries, whose inferred ontological facts are manually verified as evaluation labels). 
Since gold answers are unavailabel, labels for all evaluated items are generated by an LLM judge and manually verified by human annotators.
Table~\ref{tab:onto_construction_quality_cost} reports the construction quality under two settings: 
(1) the w/o LLM setting removes LLM from ontology-view KG construction. 
Entity linking adopts the top ReFinED candidate, falling back to the top Wikidata API candidate when ReFinED returns no candidate. For unlinked entities, the retrieval model selects the top 6 concepts, followed by relation assignment as described in Section~\ref{subsec:3.1};
(2) the w/ LLM setting follows the full construction, where the LLM selects Wikidata entities from retrieved candidates and ranks concept candidates for unlinked entities before relation assignment.
For entity linking, the w/ LLM setting maintains a high $Link$ and higher EL-Acc across all data sets, indicating that most entities are correctly linked to Wikidata, while the w/o LLM setting gives a larger $Link$ with lower EL-Acc, indicating more incorrect matches.
For linked entities, both settings obtain reasonable L-MiF1/L-MaF1 (with better overall performance under w/ LLM), showing that our method remains effective with real-world ontological facts. 
For unlinked entities, our pipeline achieves effective entity typing under w/ LLM (U-MiF1/U-MaF1 substantially exceed the scores under w/o LLM), confirming its effectiveness for concept assignment.


\noindent\textbf{Ontology Construction Cost.}
Table~\ref{tab:onto_construction_quality_cost} reports 5 indicators for ontology-view KG construction: $Ent$ (number of TKG entities), $Time$ (total construction time), $LLM$ (number of LLM calls), $API$ (external requests to Wikidata API and Wikipedia API), and $SPARQL$ (SPARQL endpoints). 
Results confirm that time scales approximately linearly with $Ent$, consistent with our complexity analysis.
Using an LLM requires more time but yields higher quality, demonstrating that our construction pipeline is both practical and reliable.

\noindent\textbf{Effect of Different LLMs and Retrieval Models.}
We compare $3$ LLMs (i.e., GPT-4o-mini~\cite{gpt4}, GPT-5-mini~\cite{gpt5}, GPT-5-nano~\cite{gpt5}) and $3$ retrieval models (i.e., Qwen3-Embedding-0.6B~\cite{qwen3-embedding}, mxbai-embed-large-v1~\cite{mxbai_embed}, and bge-small-en-v1.5~\cite{bge}) in terms of L-MaF1 and U-MaF1 on ICEWS14/ICEWS18 (Figure~\ref{fig:model_choice}). We can see that stronger models generally improve typing quality and entity coverage. 
GPT-4o-mini with Qwen3-Embedding-0.6B, which we use as default, achieves competitive performance on both data sets.

\begin{table}[!t]
\renewcommand{\arraystretch}{0.55}
\centering
\caption{OntoTKGE runtime (s/epoch) on 5 data sets; red arrows indicate overhead vs. base model.}
\vspace{-4mm}
\setlength{\tabcolsep}{1.6pt}
\newcommand{\timecost}[2]{#1\,{\scriptsize\textcolor{red}{$\uparrow$#2}}}
\begingroup
\begin{tikzpicture}
\node[
    inner sep=1.5pt,
    outer sep=1.5pt
] (tableimage) {
\scalebox{0.76}{
\begin{tabular}{llccccc}
\hline
\multirow{2}{*}{\emph{\textbf{Base model}}} & \multirow{2}{*}{\emph{\textbf{Phase}}} & \multicolumn{5}{c}{\emph{\textbf{Data set}}} \\
\cline{3-7}
 & & \emph{ICEWS14} & \emph{ICEWS18} & \emph{ICEWS05-15} & \emph{GDELT} & \emph{Wiki35k} \\
\hline
\multirow{2}{*}{RE-GCN} & Train & \timecost{19}{2} & \timecost{57}{4} & \timecost{597}{29} & \timecost{117}{5}  & \timecost{38}{9} \\
 & Test & \timecost{3}{0} & \timecost{18}{0} & \timecost{66}{1} & \timecost{88}{2}  & \timecost{20}{1} \\
\hline
\multirow{2}{*}{TiRGN} & Train & \timecost{115}{2} & \timecost{636}{56} & \timecost{2129}{11} & \timecost{1215}{3}  & \timecost{824}{11} \\
 & Test & \timecost{11}{1} & \timecost{105}{3} & \timecost{228}{4} & \timecost{293}{5}  & \timecost{80}{1} \\
\hline
\multirow{2}{*}{RETIA} & Train & \timecost{276}{19} & \timecost{117}{7} & \timecost{2855}{24} & \timecost{487}{7}  & \timecost{732}{87} \\
 & Test & \timecost{16}{1} & \timecost{22}{1} & \timecost{219}{7} & \timecost{134}{6}  & \timecost{38}{3} \\
\hline
\multirow{2}{*}{LogCL} & Train & \timecost{89}{4} & \timecost{186}{6} & \timecost{1146}{111} & \timecost{1097}{13}  & \timecost{264}{6} \\
 & Test & \timecost{8}{0} & \timecost{36}{3} & \timecost{166}{2} & \timecost{457}{7}  & \timecost{31}{1} \\
\hline
\multirow{2}{*}{HisRES} & Train & \timecost{378}{8} & \timecost{2096}{11} & \timecost{5059}{12} & \timecost{9786}{45}  & \timecost{785}{52} \\
 & Test & \timecost{42}{2} & \timecost{301}{4} & \timecost{641}{6} & \timecost{1269}{18}  & \timecost{78}{2} \\
\hline
\end{tabular}
}
};
\end{tikzpicture}
\endgroup
\vspace{-4.5mm}
\label{tab:time}
\end{table}


\begin{figure}[!t]
  \centering
    \begin{tikzpicture}
      \node[
        inner sep=1pt,
        outer sep=1pt
      ] (image) {
          \includegraphics[width=0.84\linewidth]{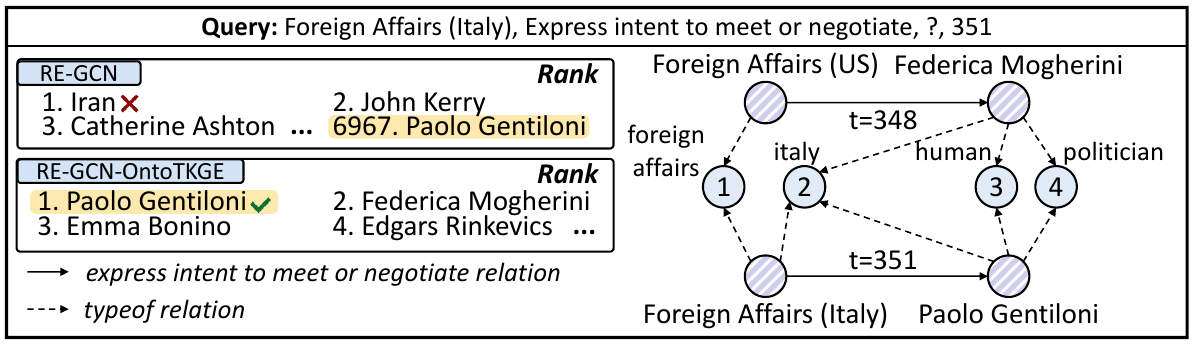}
      };
    \end{tikzpicture}
  \vspace{-4mm}
  \caption{Case study.}
  \vspace{-6mm}
  \label{fig:case_study}
\end{figure}

\vspace{-2mm}
\subsection{Efficiency Study}
To evaluate the efficiency of OntoTKGE, we test the execution time (i.e., training time and testing time) for OntoTKGE equipped with different TKG extrapolation models and their corresponding base models on various data sets. Training and inference times are evaluated in seconds for consistent comparison. 
The results in Table \ref{tab:time} show that OntoTKGE brings a slight overhead to training and inference times. 
Note that the increase is marginal across all data sets, demonstrating the efficiency of OntoTKGE. 

\vspace{-2mm}
\subsection{Case Study}
Figure \ref{fig:case_study} shows a typical case from the ICEWS14's test set. We compare RE-GCN-OntoTKGE with its base model RE-GCN on the query (\textit{Foreign Affairs (Italy)}, \textit{Express intent to meet or negotiate}, ?). The result shows that RE-GCN ranks the gold tail entity \textit{``Paolo Gentiloni''} at rank 6967, whereas RE-GCN-OntoTKGE successfully ranks it first. As shown in the right of Figure \ref{fig:case_study}, this gain suggests that our framework OntoTKGE can capture relevant ontological knowledge from similar historical events (e.g., the fact at $t=348$) via shared concepts, thereby inferring the sparse entity.

\vspace{-3mm}
\section{Related Work}
\noindent\textbf{TKG Extrapolation.} 
KG evolution refers to structural and content changes in KGs over time. Within KG evolution, TKGs focus on the temporal validity of facts, differing from versioned KGs (full versions of the KG at different points in time) and dynamic KGs (continuous edge/node‑level changes)~\cite{evo1}. Here, we focus on TKG extrapolation.
Early studies relied on historical snapshots. 
RE-GCN \cite{regcn} learns evolutional representations from local historical dependency, and TiRGN \cite{tirgn} extends it with local-global historical patterns via a local recurrent encoder with a global history encoder.
Recently, RETIA \cite{retia} introduced line graphs for better relation interactions and neighborhood aggregation. 
LogCL \cite{logcl} integrated temporal evolution and global repetition historical patterns via contrastive learning.
HisRES \cite{hisres} models recent inter-snapshot correlations and long-term dependencies with carefully designed encoders.
Yet, learning embeddings for sparse entities remains a challenge. ANEL \cite{anel} addresses this by incorporating neighbor information, providing greater support to sparse entities. 
LLM-DA \cite{llmda} extracted and dynamically updated rules via LLMs, integrating rule-based and extrapolation models for prediction.
Unlike these methods relying on heuristics from KG snapshots, OntoTKGE solves this by utilizing explicit ontological knowledge from an ontology-view KG. 

\noindent\textbf{Ontology-Enhanced KG Embedding.}
Ontological concepts are increasingly used to improve entity embeddings. JOIE \cite{joie} pioneered this direction by proposing a universal framework to jointly learn embeddings from intra-view relations and cross-view links.  
CISS \cite{ciss} enhanced embeddings by jointly modeling class inheritance and structural similarity, capturing fine-grained hierarchical semantics. 
HyperCL \cite{hypercl} separately learned embeddings from the instance-view KG and the hierarchical ontology-view KG, and aligned them via a concept-aware contrastive loss to enforce semantic consistency. 
OntoTKGE makes the first attempt to enhance TKG extrapolation through the use of ontological knowledge. OntoTKGE is an encoder-decoder framework that combines a global ontology-aware evolutionary encoder with a local ontology-aware relevance encoder.
Ontology-enhanced static KG completion methods~\cite{joie,ciss,hypercl} (adapted to TKG extrapolation) and OntoTKGE's global encoder both enhance the entity embeddings of the same TKG base model with ontological knowledge, yet differ in mechanism: the former employ alignment losses to constrain the embeddings, while the latter incorporates ontological knowledge to initialize entity embeddings.
Beyond that, OntoTKGE uses a local encoder to generate embeddings for query entities from their corresponding ontology-view subgraphs, alleviating ontological knowledge loss across KG snapshots, and fuses them with global encoder outputs via a contrastive-enhanced gated fusion unit. This design fundamentally differs from existing ontology-enhanced KG embedding methods, making OntoTKGE better suited for TKG.
Additionally, existing ontology-enhanced static KG completion methods focus on representation learning, and they largely overlook ontology reuse. Reused ontologies evolve asynchronously and independently, and changes in reused terms affect downstream reuse~\cite{evo1,evo2}. 
OntoTKGE mitigates this by reusing the construction cache to update the ontology-view KG with the latest external ontology.

\vspace{-4mm}
\section{Conclusion and Future Work}
We propose OntoTKGE, a framework for TKG extrapolation that handles sparse entities by integrating ontological and temporal knowledge effectively. It is flexible to adapt to many TKG extrapolation models. 
Experiments on $5$ data sets show it improves many SOTA TKG extrapolation models and surpasses all baselines. 
Future work will include a deeper investigation of the use of Conflict and Mediation Event Observations (CAMEO) in OntoTKGE, more capable LLMs for higher construction quality and better performance of TKG extrapolation, and mechanisms for updating ontology-view KGs as external ontologies evolve.

\vspace{-4mm}
\begin{acks}
The work was partially supported by the National Key Research and Development Program of China (No. 2024YFF0617702); the National Natural Science Foundation of China (Nos. U22A2025, 62402097, 62232007, and U23A20309); the Joint Funds of the Natural Science Foundation of Liaoning Province (No. 2023-BSBA-132); the 111 Project (No. B16009); 
and the Fundamental Research Funds for the Central Universities (No. N2417007).
\end{acks}


\bibliographystyle{ACM-Reference-Format}
\bibliography{sample}

\end{document}